\documentclass{article}

    \usepackage[preprint]{neurips_2025}

\usepackage[utf8]{inputenc} 
\usepackage[T1]{fontenc}    
\usepackage{hyperref}       
\usepackage{url}            
\usepackage{booktabs}       
\usepackage{amsfonts}       
\usepackage{nicefrac}       
\usepackage{microtype}      
\usepackage{xcolor}         
\usepackage{amsmath}
\usepackage{graphicx}
\usepackage{multirow}
\usepackage{algorithm}
\usepackage{algpseudocode}

\title{multivariateGPT: a decoder-only transformer for multivariate categorical and numeric data}

\author{%
  Andrew J. Loza$^{1,2}$ \thanks{Code available at: 
\url{https://anonymous.4open.science/r/multivariateGPT_anon-4ED4/README.md}}
  \And
  Jun Yup Kim$^4$
  \And
  Shangzheng Song$^4$ 
  \And
  Yihang Liu$^4$ 
  \AND
  Joseph J. Y. Sung$^4$ 
  \And
  R Andrew Taylor$^4$ 
  \And
  Dennis L. Shung$^3$ 
  \AND
  \normalfont
  $^1$Department of Biomedical Informatics and Data Science, Yale School of Medicine\\
  $^2$ Department of Pediatrics, Yale School of Medicine\\
  $^3$ Department of Medicine, Yale School of Medicine\\
  $^4$ Yale School of Medicine\\
  $^4$ Lee Kong Chian School of Medicine, Nanyang Technological University
}

\begin{document}

\maketitle

\begin{abstract}
Real-world processes often generate data that are a mix of categorical and numeric values that are recorded at irregular and informative intervals. Discrete token-based approaches are limited in numeric representation capacity while methods like neural ordinary differential equations are not well suited for categorical data or informative sampling and require augmentation to handle certain classes of trajectories. Here, we present multivariateGPT, a single architecture for modeling sequences of mixed categorical (including tokenized text) and numeric data. This is accomplished with an autoregressive sequence decomposition, embedding scheme, and loss function that extend the next token prediction task to likelihood estimation of the joint distribution of next token class and value. We demonstrate how this approach can efficiently learn to generalize patterns in simple physical systems and model complex time series including electrocardiograms and multivariate electronic health record data. This work extends the utility of transformer based models to additional classes of data.
\end{abstract}

\section{Introduction}
\label{introduction}

Data collected in complex systems are often a mixture of categorical and numeric values that are informatively and sparsely sampled across a range of classes. Each of these features presents technical challenges, and current approaches are often limited in modeling one or more of these aspects \citep{shukla2020survey}.

One class of models are autoregressive neural network-based sequence models, such as recurrent neural networks, long-short term memory networks \citep{hochreiter1997long}, and transformers \citep{vaswani2017attention}. However, these models do not natively handle categorical data (including as text) and numeric data within a single model. 

Using language models as a starting point, four common approaches to create a multimodal model that processes numeric and categorical data simultaneously are (a) tokenization of strings representing numeric quantities \citep{singh2024tokenization, gruver2023large} (b)  alternate embedding strategy for numeric tokens such as XVal or MMD \citep{zausinger2024regress,alberts2024interleaving}(c) discretization of numeric values \citep{zhu2024prompting} and (d) shared embedding space with a model designed for the specific type of numeric data used, such as with image data \citep{sun2023emu}. The first three approaches operate within the existing language modeling framework. However, numeric data from different classes shares a common representation and rely on the context of the preceding token to differentiate meaning. Approaches (a) and (c) both rely on categorical representation of numeric quantities preventing shared information across related tokens: if tokens representing values of $21$ and $23$ are seen in training, there is no information shared with the token representing $22$. Approach (d) introduces additional complexity of training a separate model for the numeric domain or having domain specific encoding and decoding methods. Additional challenges occur when numeric data modalities can have variable size, often requiring resizing or alternative methods before encoding \citep{han2022survey,tang2025data}.

A second class of models are based on neural ordinary differential equations which use a continuous representation of the timeseries \citep{chen2018neural}. These models handle irregular sampling but cannot represent crossing trajectories. Augmentation addresses this issue \citep{rubanova2019latent}, but cannot handle stochasticity without conversion to a stochastic differential equation approach which increases complexity\citep{kidger2021neural}. Trajectory flow matching has improved this through simulation-free training\citep{zhang2024trajectory}. However, these models are limited in modeling categorical data and extracting information from sampling measurement timing.

A concrete example of the critical importance of models with these capabilities is in healthcare. Data includes categorical values such as diagnostic codes or text as well as numeric data such as laboratory values, vital signs, and medication dosages. There are thousands of potential classes of data, but only a few are measured at any given time. Additionally, the timing of observations reflect a decision to record an observation, which itself contains information \citep{getzen2023leveraging}. 

Here we present multivariateGPT, a decoder-only transformer that uses a single architecture to model mixed sequences of categorical and numeric data and has a likelihood-based loss function. This is a multimodal model; however, we refer to it as \textit{multivariate} because all data is handled within a single model architecture without linking two models. We offer three main contributions:

\begin{enumerate}
    \item We provide an auto-regressive decomposition for the joint distribution of multivariate time-series which captures information on the timing of measurements which can be approximated by a decoder-only transformer.
    \item We present tokenization, embedding, and loss methods that allow for likelihood-based next token prediction for both categorical (including text) and numeric values in a single architecture. This method is compatible with a pure language modeling objective, such as if a "measurement" in the time series is a document. The loss function also allows for uncertainty estimation of numeric quantities.
    \item We provide an implementation of this model and demonstrate its ability to (a) model second-order dynamics physical systems, generalizing to predict trajectories outside of its training data (b) model multivariate clinical data from electronic health records, and (c) model electrocardiogram timeseries which have stochasticity and substantial nonlinearities.
\end{enumerate}

\section{Methods}
\label{methods}

\subsection{Multivariate Generative Modeling}
We consider the task of modeling the joint distribution of data collected in a dynamic system. These variables may be categorical (including tokenized text) or numeric and are recorded at specific times. They may represent measurements of the system or actions taken on the system. The joint distribution can be decomposed via the chain rule of probability into an auto-regressive sequence prediction task which uses the natural sequential ordering of the data elements:
\begin{equation}\label{eq:1}
    P(\{\bold{X},\tau\}_0^i) = \prod_{i=1}^n P(\{\bold{X},\tau\}_i | \{\bold{X},\tau\}_1, ... , \{\bold{X},\tau\}_{i-1})
\end{equation}
where $\bold{X}$ is a set of observations recorded with the same time-stamp and $\tau$ is the elapsed time since the prior data record. Each $\{\bold{X},\tau\}$ tuple represents a joint distribution over the specific events belonging to $\bold{X}$ and the value of $\tau$. Each term on the right-hand-side of Eq. \ref{eq:1} can be further decomposed:
\begin{equation}\label{eq:2}
    P(\{\bold{X},\tau\}_i | \bold{S}_{i-1}) = P(\tau_i|\bold{S}_{i-1})\prod_{j=1}^m P(x_j, |x_1, ..., x_{j-1},\tau_i, \bold{S}_{i-1})
\end{equation}
Where $\bold{S}_{i-1}$ is shorthand for the the prior sequence of $\{\bold{X},\tau\}$ tuples, $m$ is the number of data elements in the set $\bold{X}$, and $x_j$ are individual data elements within $\bold{X}$.

In this multivariate framework, each $x_j$ variable represents a tuple of class and value such that $x_j = (c_j,v_j)$ for $c_i \in \bold{C}$ and value $v_i \in V_{c_i}$, where $\bold{C}$ are all possible classes of recorded data and $V_{c_i}$ are categorical or numeric values specific to class $c_i$. Concrete examples from healthcare data are: numeric $(creatinine,1.21)$, categorical $(diagnosis,Z00.1)$, text: $[(text,the), (text,patient), (text,is)]$.
A specific ordering can be imposed over the $x_j$ within $\bold{X}$ which respects the required sequence order (such as for text data) and otherwise uses a lexicographic sort over classes. This is similar to the pixel channel ordering in autoregressive image models.

By framing the wait time $\tau$ as its own class and value, we can rewrite Eq. \ref{eq:2} as a completely flattened autoregressive sequence of conditional probabilities:
\begin{align}
        P(\bold{X}) &= \prod_{i=1}^k P(x_i, |x_1, ..., x_{i-1}), \quad
        \text{or expanded:}\\ 
        P(\bold{X}) &= \prod_{i=1}^k P(\{c_i,v_i\} |\{c_1,v_1\}, ..., \{c_{i-1},v_{i-1}\})
\end{align}
where $\bold{X}$ is now the full sequence of all $x_i = \{c_i,v_i\}$ tuples including the time class. As a secondary note, $(c_i,v_i)$ tuples where $v_i$ is categorical, will be transformed into a set of new classes where each new class is the combination of the original $c_i$ and each level of  $V_{c_i}$. This expands the number of classes, but enforces that each categorical class has a single value which will allow for a simpler model design. This decomposition can then be estimated by any autoregressive sequence model which is modified to predict the joint distribution of $P(c_i,v_i)$ for any prior input. We next describe an embedding scheme and loss function to achieve this with a decoder-only transformer.

\subsection{Embeddings}
Each tuple $\{c_i,v_i\}$ is considered a token and mapped to an embedding vector $E_i \in \mathbb{R}^e$ where $e$ is the embedding dimension as follows:
\begin{equation} \label{eq:5}
\\
E_i = 
\begin{cases}
E_{c_i}+f_{c_i}(v_i) &\text{ for }v_i \in \mathbb{R}^n\\
E_{c_i} &\text{ for categorical }v_i
\end{cases}
\end{equation}
Where $E_{c_i}$ are class-specific learned embedding vectors, and $f_{c_i}(v_i)$ is any function mapping $\mathbb{R}^n$ to $\mathbb{R}^e$. Note that because categorical classes are re-defined to contain a single value, only a single embedding vector is needed for each.

\subsection{Loss function}

The sequence of tokens is modeled in an autoregressive fashion using a decoder-only generative pretrained transformer architecture. To maintain a single model without the use of a contrastive loss, we construct an output and loss function that allows for minimizing the negative log likelihood of the joint distribution $P(c_i,v_i)$ of the next token:
\begin{equation}
    P(c_i,v_i) = P(c_i)P(v_i|c_i)
\end{equation}

The token embedding in the final layer of the model is projected via two heads to estimate $P(c_i)$ and $P(v_i|c_i)$. The first head is equivalent to the standard language model head and projects $\mathbb{R}^{d_e} \rightarrow \mathbb{R}^{d_c}$ followed by a \textit{softmax} function to estimate the likelihood of the next token across each class. The second head is used to estimate $P(v_i|c_i)$ and could take multiple forms; here we use a projection from $\mathbb{R}^{d_e} \rightarrow \mathbb{R}^{d_c\times2}$ where the output is used as $\mu$ and $\sigma$ of normal distribution using \textit{softplus} to ensure positive $\sigma$. The per-sample loss can then be defined as a joint log-likelihood:
\begin{align}
    l_{ci} = -\sum_{j=1}^Cc_{j}\log(\hat{c}_{j}), \quad 
    l_{vi} = -\sum_{j=1}^Cc_{j}\log P(v_{j}|\hat{\mu}_{j},\hat{\sigma}_{j})
\end{align}
where $l_{ci}$ is the per-token class loss, $l_{vi}$ is the per-token conditional value loss, $c_{j}$ is a binary indicator of the correct class index, $\hat{c}_{j}$ are predicted probabilities across classes, $v_{j}$ is a vector with one non-zero element equal to the correct value at the index of the correct class and $\hat{\mu}_{j}$, $\hat{\sigma}_{j}$ are the predicted mean and variance for value of the $j^{th}$ class.  Categorical classes have only one value, such that $P(v|c) = 1$ and  $l_{vi} = 0$. Alternatively, a constant $w$ can be used as weighting. The batch loss is defined as:
\begin{equation} \label{eq:9}
        \mathcal{L} = \frac{1}{N}\sum_{i=1}^{N} l_{ci}+l_{vi}
\end{equation}

\subsection{Algorithm}
This method is summarized in Alg. \ref{alg:mgpt}. This assumes that time deltas between $x^{(i)}$ have been converted to class-value tuples $(c\gets time,v \gets t^{(i)}-t^{(i-1)})$ for $t^{(i)}-t^{(i-1)} > 0$ in data preprocessing.

\begin{algorithm}[h]
\small
\caption{Mutivariate autoregressive model}
\label{alg:mgpt}
\begin{algorithmic}
\Require Data $\mathcal{D} = \{x^{(i)}\}_{i=1}^N$, where $x^{(i)} = [(c_1, v_1), \dots, (c_k, v_k)]$
\Require Model $f_\theta$ with parameters $\theta$
\While{training}
    \State $x^{(j)} \sim \mathcal{U}(\mathcal{D})$
    \State $\mathcal{L} \gets 0$
        \For{$t = 2$ to $k$}
            \State $(\hat{p}_{\text{class}}, \hat{\mu}_{c}, \hat{\sigma}_{c}) \gets f_\theta(x^{(1:t-1)})$
            \State $L_{\text{class}} \gets -\log \hat{p}_{\text{class}}[c_t]$
            \State $L_{\text{value}} \gets -\log \mathcal{N}(v_t; \mu_c, \sigma_c)$
            \State $\mathcal{L} \gets \mathcal{L} + \mathcal{L}_{\text{class}} + \mathcal{L}_{\text{value}}$
        \EndFor

        \State $\theta \gets \mathrm{Update}(\theta,\nabla_\theta \mathcal{L})$

\EndWhile
\State \Return $f_\theta$
\end{algorithmic}
\end{algorithm}

\subsection{Model Architecture}
\begin{figure*}[h]
    \centering
    \includegraphics[width=0.9\textwidth]{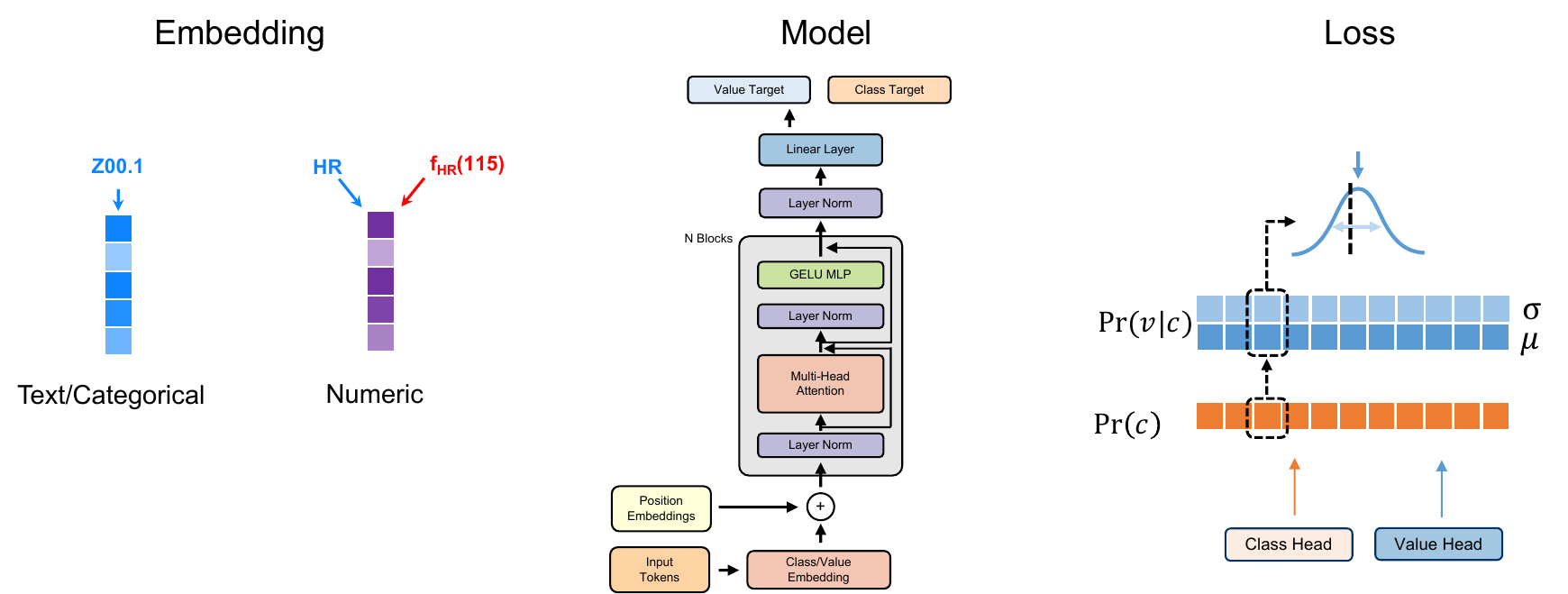}
    \caption{Diagram of Model Architecture.}
    \label{fig:diagram}
\end{figure*}

We use a decoder-only transformer to approximate Eq. \ref{eq:1} by minimizing Eq. \ref{eq:9}. The model architecture follows the nanoGPT implementation of a decoder-only transformer \citep{Karpathy2022} with modifications to the embedding scheme and loss function to implement methods described in Section \ref{methods}. A schematic is shown in Figure \ref{fig:diagram}. In this implementation, a single feed-forward layer is used for value mapping. The final linear layer consists of a Class Head and Value Head which are used to estimate the loss based on the joint probability of the next token's class and value.

\section{Experimental Results}

Experiments were run on a computing cluster with a heterogenous cluster of NVIDIA RTX4080, RTX5000 and A40 GPUs for approximately 5,000 GPU hours. Individual training and inference runs require approximately one GPU day.

\subsection{Fitting and Generalization in Simple Harmonic Oscillators}
Trajectories of damped simple harmonic oscillators provide a toy system that requires the ability to model second order dynamics and crossing trajectories. Neural ordinary differential equation models, canonical conditional flow matching, and bridge matching cannot fit these trajectories due to the presence of crossing trajectories \citep{zhang2024trajectory}. We compared the multivariateGPT model to discrete token transformer baselines on two tasks: trajectory reconstruction and generalization. 

\begin{figure*}[h]
    \centering
    \includegraphics[width=0.9\textwidth]{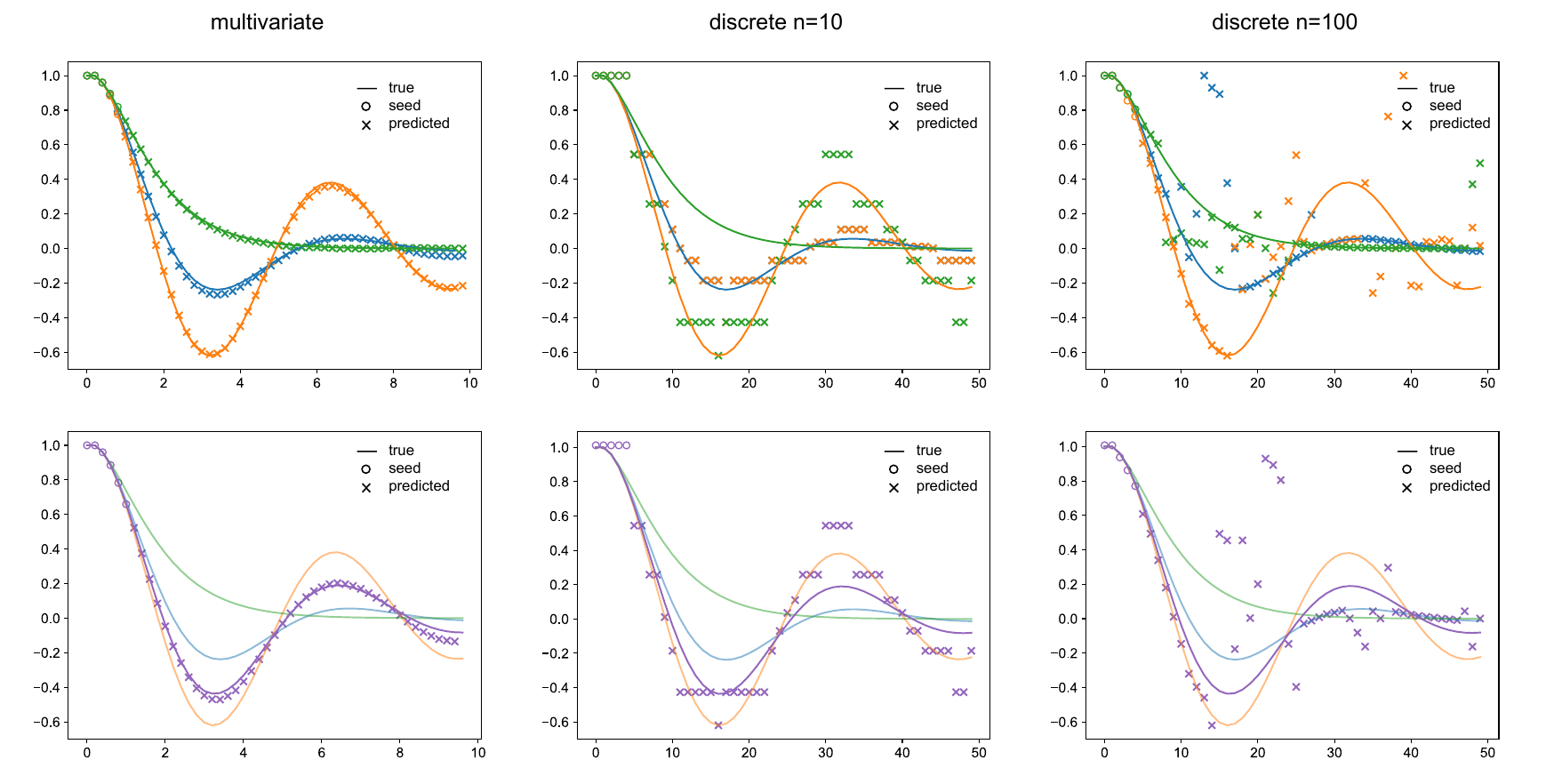}
    \caption{Simple harmonic oscillator trajectories. Columns show results from each model type. First Row: reconstruction of training data. Second Row: generalization to unseen trajectories.}
    \label{fig:oscillator}
\end{figure*}

\textbf{Trajectory reconstruction:} The first task was the reconstruction of trajectories from the training data using the seed of the first 5 points (Fig. \ref{fig:oscillator}, top row). The multivariate model reconstructs trajectories without accumulating error. The discrete transformer model with 10 bins lacks the numeric precision to resolve trajectories from the first 5 points, resulting in reproduction of a random trajectory from the training data. With 100 bins, numeric precision is high enough to differentiate trajectories from the first 5 points, however, reconstructions suffer from significant noise accumulation.

\textbf{Trajectory generalization:} In the second task, the models are given the first 5 points from a trajectory that was not present in the training data. The multivariate model generalizes to this unseen trajectory. The discrete model with 10 bins does not generalize and reproduces the orange trajectory from the training data. The discrete model with 100 bins initially approximates the orange trajectory and then further degrades as noise accumulates. Increasing training time by a factor of 10 improves the fit of the discrete model with 100 bins, but generalization still fails (Appendix \ref{appendix:oscillator}). Ablation studies demonstrate that allowing the model to learn variance is important for precise next-token value estimates that allow for reconstruction with minimal error and generalization (Appendix \ref{appendix:oscillator}).

\subsection{Clinical Data sets}
We evaluated the performance of multivariateGPT on real-world clinical data sets that encompass a range of time-series characteristics.

\begin{enumerate}
    \item \textbf{eICU Sepsis Data:} subset of patients in the eICU Collaborative Research Database v2.0 with a diagnosis of sepsis, chosen to match \citep{zhang2024trajectory}. Includes 3,362 patients, with data on age, sex, heart rate, mean arterial pressure (MAP), and norepinephrine infusion rate during the first 24 hours of ICU admission. This was derived from the eICU collaborative database \citep{pollard2018eicu}.
    \item \textbf{MIMIC-IV Electrocardiogram Data:} subset of 89,339 electrocardiograms from the MIMIC-IV electrocardiogram database. Each record includes patient age, sex, and voltages for the 12 standard leads of an electrocardiogram. This was taken from MIMIC-IV-ECG dataset \citep{gow2023mimic} derived from the MIMIC-IV database \citep{johnson2023mimiciv}. 
    \item \textbf{Physionet ICU Data:} data on 65,155 patients admitted to the the ICU from 3 hospital systems with demographics and hourly records of vital signs and lab values representing 36 unique classes \citep{moor2021early}.
\end{enumerate}

\subsubsection{eICU: Heart Rate and Blood Pressure Prediction in Sepsis}
We evaluated the multivariateGPT model compared to discrete-token transformer baselines, an open-source LLM, and TFM-ODE, which recently outperformed a range of neural ODE and SDE methods in this data set.  This data set has irregularly and potentially informatively sampled heart rate and MAP measurements. As in \citep{zhang2024trajectory}, we focused on a subset of patients admitted with sepsis as their primary diagnosis. Two tasks were tested due to differences in what each model class was able to predict. While the multivariateGPT and discrete-token transformer models predict both measurements and the time between measurements, TFM-ODE requires specifying the time at which measurements occur.

\textbf{Value prediction:} In the first task, data from the first 3, 6, 9, and 12 hours were given as seeds, and models were evaluated on the accuracy of predicting the remaining heart rate and MAP measurements. Measurement times were provided to allow comparison to TFM-ODE and autoregressive infilling of heart rate and MAP measurements were used for the transformer models. The multivariate GPT consistently achieved the lowest MSE among all compared models with error reductions of 40\% to 60\% compared to the next best performing model, TFM-ODE (Table \ref{tab:mse_results}). Most models demonstrated improved performance with longer seed periods.

\begin{table}[ht]
\centering
\small
\caption{Mean Squared Error (± Standard Error) for MAP and Heart Rate from various seed lengths for each model.}
\begin{tabular}{lcccc}
\toprule
\textbf{Model} & \textbf{3h} & \textbf{6h} & \textbf{9h} & \textbf{12h} \\
\midrule
Gemma 1.1-7b & 0.0999$\pm$0.0050 & 0.1246$\pm$0.0094 & 0.0461$\pm$0.0018 & 0.0299$\pm$0.0016 \\
TFM\_ODE & 0.0268$\pm$0.0004 & 0.0236$\pm$0.0004 & 0.0166$\pm$0.0003 & 0.0146$\pm$0.0003 \\
Discrete (n=10) & 0.1183$\pm$0.0028 & 0.1108$\pm$0.0029 & 0.1053$\pm$0.0031 & 0.1016$\pm$0.0038 \\
Discrete (n=50) & 0.0308$\pm$0.0012 & 0.0295$\pm$0.0011 & 0.0279$\pm$0.0014 & 0.0276$\pm$0.0017 \\
Multivariate GPT & \textbf{0.0108$\pm$0.0002} & \textbf{0.0104$\pm$0.0003} & \textbf{0.0098$\pm$0.0003} & \textbf{0.0087$\pm$0.002} \\
\bottomrule
\end{tabular}

\label{tab:mse_results}
\end{table}

\textbf{Value and timing prediction:} In the second task, we evaluated the performance of the multivariate GPT model in predicting both the time and value of future observations. This ability is important in medicine because the timing of a measurement contains information about the state of a patient. This feature cannot be modeled by current neural ODE models or TFM-ODE. In this task, models were provided a seed of the first 3 hours, 6 hours, 9 hours, and 12 hours of data for a patient trajectory. The multivariate GPT model outperforms the discrete transformer model across all time windows and in both the value and time prediction task  (Table \ref{tab:value_time_mse}). Error reduction for the multivariateGPT model ranged from 82\% to 97\% in value prediction and 78\% to 99\% in time prediction.

\begin{table}[ht]
\centering
\small
\caption{Mean squared error (± Standard Deviation) for value and time predictions. TFM-ODE is not included because it does not predict the timing of measurements.}
\small
\begin{tabular}{lcccc}
\toprule
\textbf{Model} & \textbf{3h} & \textbf{6h} & \textbf{9h} & \textbf{12h} \\
\midrule
\multicolumn{5}{l}{\textbf{Value Prediction (MSE)}} \\
Discrete n10        & 0.3892 ± 0.0745 & 0.2598 ± 0.0390 & 0.2581 ± 0.0399 & 0.3924 ± 0.0155 \\
Discrete n50        & 0.0763 ± 0.0293 & 0.0560 ± 0.0161 & 0.0530 ± 0.0142 & 0.0717 ± 0.0355 \\
Multivariate GPT    & \textbf{0.0114 ± 0.0013} & \textbf{0.0082 ± 0.0017} & \textbf{0.0078 ± 0.0016} & \textbf{0.0126 ± 0.0013} \\
\midrule
\multicolumn{5}{l}{\textbf{Time Prediction (MSE)}} \\
Discrete n10        & 6.8673 ± 2.4835 & 4.1782 ± 0.2968 & 6.2214 ± 0.3843 & 7.9618 ± 0.3171 \\
Discrete n50        & 0.1783 ± 0.0787 & 0.4365 ± 0.0926 & 0.8600 ± 0.3455 & 1.4893 ± 0.4038 \\
Multivariate GPT    & \textbf{0.0376 ± 0.0051} & \textbf{0.0230 ± 0.0008} & \textbf{0.0279 ± 0.0020} & \textbf{0.0341 ± 0.0035} \\
\bottomrule
\end{tabular}
\label{tab:value_time_mse}
\end{table}

\begin{figure*}[h]
    \centering
    \includegraphics[width=0.9\textwidth]{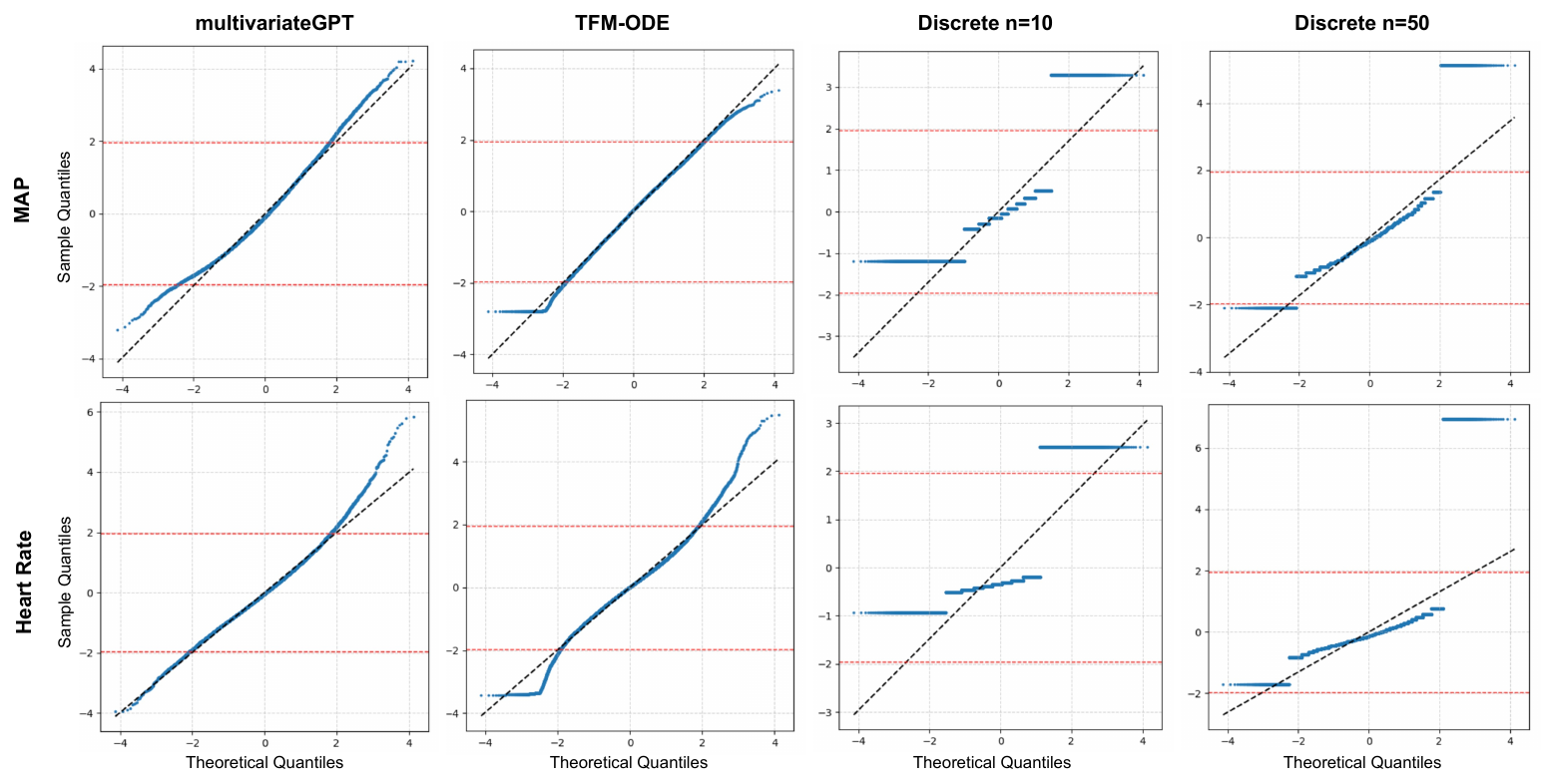}
    \caption{QQ Plots. Theoretical quantiles are plotted against the sample quantiles for each model (columns). The top row are plots for MAP predictions and the bottom row are plots for heart rate predictions. Dashed bands are located at z-scores of $-1.96$ and $+1.96$.}
    \label{fig:qqplot}
\end{figure*}

\textbf{Calibration:} The multivariateGPT model and TFM-ODE model generate both expected value and variance for each observation. The calibration of the estimated variance is shown in quantile-quantile (QQ) plots comparing the standardized residuals (z-scores) of the true value to the theoretical values from a normal distribution Fig. \ref{fig:qqplot}. The multivariateGPT model and TFM-ODE are well calibrated for central theoretical quantiles while the discrete models show marked departure from linearity. Compared to TFM-ODE, the multivariateGPT model shows improved calibration for the leftmost quantiles in heart rate estimation. Coverage fractions for predicted 95\% confidence intervals, defined as the fraction of the true values falling within predicted $\mu \pm1.96*\sigma$ intervals, are shown in Table \ref{tab:coverage}. 

\begin{table}[h]
\centering
\small
\caption{Empiric coverage fractions for predicted 95\% confidence intervals. Values closer to 0.95 correspond to better calibration. Values are MSE $\pm$ standard deviation.}
\begin{tabular}{lcc}
\toprule
\textbf{Method} & \textbf{MAP} & \textbf{Heart Rate} \\
\hline
multivariateGPT & \textbf{0.951 $\pm$0.009} & \textbf{0.957$\pm$0.005} \\
TFM-ODE & 0.932$\pm$0.066 & 0.966$\pm$0.022\\
Discrete (n=10) & 1.000$\pm$0.000* & 1.000$\pm$0.000* \\
Discrete (n=50) & 0.9983$\pm$0.0004 & 0.9872$\pm$0.0130 \\
\hline
\end{tabular}
\label{tab:coverage}
\begin{center}
*Due to the bin sizes, all values are positioned within the central 95\% of the distribution.
\end{center}
\end{table}

\subsubsection{MIMIC-IV Electrocardiogram Data}

Electrocardiogram signals are quasi-periodic, noisy, and have significant nonlinear behavior, which creates unique challenges that are separate from the eICU data above. Data were decomposed into a sequence of tokens: $
 [(\text{Age}, 63), (\text{Sex}, \text{Male}), (\text{Lead I}, 0.12), (\text{Lead II}, 0.15),$ $ \dots, $ $(\text{Lead I}, -0.03), (\text{Lead II}, 0.01), \dots]. $
 We evaluated the performance of multivariateGPT compared to a discrete transformer approach on this data set. Both classes of models were trained on the data set and tested on the task of lead reconstruction. The model is provided with the ground truth for a subset of leads (here I, II, V2, V6), and the model is tasked with autoregressivelly infilling the masked leads (8 remaining leads). The multivarateGPT model performs substantially better across all lead reconstruction tasks compared with the discrete models (Table \ref{tab:leads}). We find mixed performance when comparing the discrete model with 50 bins to the model with 100 bins suggesting tradeoffs in precision and ability to learn representations for a larger vocabulary. In general performance on the limb lead reconstruction is better than on precordial leads which is often seen due to the depolarization vectors that each lead represents.
 
\begin{table}[h]
\centering
\small
\caption{Limb leads (III, aVF, aVL, aVR) values $\times 10^{-4}$, precordial leads (V1,V3,V4,V5) values $\times 10^{-2}$). Error in reconstructing full-duration trajectories for each lead.}
\begin{tabular}{lcccc}
\toprule
\textbf{Model} & \textbf{III} & \textbf{AVF} & \textbf{AVL} & \textbf{AVR} \\
\midrule
Multivariate 
& \textbf{1.94 $\pm$ 0.59} & \textbf{1.25 $\pm$ 0.41} & \textbf{1.86 $\pm$ 0.38} & \textbf{0.39 $\pm$ 0.09} \\
Discrete (n=50) 
& $106.1 \pm 32.5$ & $725.7 \pm 168.0 $ & $736.7 \pm 229.4$ & $227.5 \pm 55.9$ \\
Discrete (n=100) 
& $93.9 \pm 23.5$ & $115.6 \pm 67.0 $ & $322.7 \pm 92.9$ & $196.0 \pm 51.5$ \\
\midrule
\textbf{Model} & \textbf{V1} & \textbf{V3} & \textbf{V4} & \textbf{V5} \\
\midrule
Multivariate 
& \textbf{2.06 $\pm$ 0.41} & \textbf{4.21 $\pm$ 0.19} & \textbf{5.09 $\pm$ 1.44} & \textbf{31.0 $\pm$ 1.55} \\
Discrete (n=50) 
& $22.2 \pm 9.77$ & $38.4 \pm 28.85$ & $26.0 \pm 15.75$ & $32.0 \pm21.4$ \\
Discrete (n=100) 
& $34.6 \pm 22.3$ & $42.3 \pm 18.1$ & $22.1 \pm 13.14$ & $5.53 \pm 8.27$ \\
\bottomrule
\end{tabular}
\label{tab:leads}
\end{table}

\subsubsection{Physionet ICU Data}
The Physionet ICU data set contains measurements collected in a sparse manner across a 36 categorical and classes. This poses a unique challenge as a generative model must predict which measurements and what values will occur at any given measurement time. We evaluated the performance off multivariate GPT against discrete transformer models for next token class accuracy and next token value MSE (conditioned on correct class predictions). The multivariateGPT model performed significantly better than the discrete model with percentile based bins (Table \ref{tab:physionet}). This the discrete model erroneously predicting tokens representing extreme values. This highlights the robustness to noise and outliers that numeric embedding provides.

\begin{table}[h!]
\centering
\caption{Performance comparison of models across classification and regression metrics for Physionet ICU data. Values are $\pm$ SEM.}
\small
\begin{tabular}{lccc}
\toprule
\textbf{Model} & \textbf{Class Accuracy} & \textbf{Value MSE}  \\
\midrule
MultivariateGPT   & $0.862 \pm 0.002$   & \textbf{0.385 $\pm$ 0.018}   \\
Discrete $n=10$   & $0.863 \pm 0.002$   & $27.6 \pm 3.6$    \\
Discrete $n=50$   & $0.863 \pm 0.002$   & $10.1 \pm 0.4$    \\
\bottomrule
\end{tabular}

\label{tab:physionet}
\end{table}

\section{Related Work}
Multiple neural network-based approaches have been developed to model timeseries data to capture the complexities inherent in real-world data including mixed categorical and numeric variables, irregular or informative sampling, and stochasticity. Here we review related work for token-based and continuous function-based approaches.

\textbf{Token-Based methods}:
A naive discrete token-based approach is not optimal for numeric values \citep{spathis2024the}. Careful tokenization of strings containing digits can improve performance on tasks \citep{born2023regression}, and single digit tokenization has been found to be more efficient than tokenization into longer groups \citep{zhou2024scaling,schmidt2024tokenization}.  The LLAMA series of models adopted single-digit-based tokenization \citep{touvron2023llama} and an adaptation of these models for time series modeling enforced digit-based tokenization through addition of delimiting characters \citep{gruver2023large}. Right-to-left digit tokenization was shown to improve arithmetic performance over naive byte pair encoding strategies across a range of arithmetic tasks \citep{singh2024tokenization}. Additional efforts to transform numeric quantities into discrete tokens include discretization into bins \citep{gorishniy2022embeddings,stein2024simple,rajkomar2018scalable,renc2024zero,zhu2024prompting,huanglanguage} and a similarity-driven quantization approach for time series data \citep{zhicheng2024sdformer}. This is the main method employed in current models of medical data \citep{rajkomar2018scalable,theodorou2023synthesize,pang2024cehr,renc2024zero}. Within image and audio processing, multiple methods have been developed for transforming signals into discrete tokens including LFQ, FSQ, and BSQ which utilize vector quantization via discrete codebooks \citep{yu2023language,jia2025principles,mentzer2023finite,zhao2024image}. Embedding approaches aligned with the continuous nature of numeric values have also been developed \citep{wang2025bridging,golkar2024xval}. Once tokens are created, standard cross entropy loss is not well suited for numeric decoding and alternate loss functions including regression-like loss \citep{zausinger2024regress,han2022luna} have been developed. A remaining limitation from this existing work is a numeric representation for diverse classes of data with a single likelihood-based loss which we provide here. 

\textbf{Continuous Models}
Continuous-time trajectory models are an alternative approach to model irregularly sampled data. Models such as the latent trajectory-based Neural-ODEs \citep{rubanova2019latent, chen2018neural} and Neural-SDEs \citep{kidger2021neural,oh2024stable} have outperformed LSTM- and RNN-based methods \citep{Che2018,cao2018brits}. In turn, trajectory flow matching (TFM) outperformed these models in clinical timeseries modeling with improved computational efficiency through its simulation-free training and the additional benefit of estimating uncertainty \citep{zhang2024trajectory}. Additional methods to capture uncertainty include \citep{de_brouwer2022uncertainty}. These methods handle irregularly spaced inputs, but they do not currently derive information from the irregular spacing and do not handle categorical data well.

\section{Conclusion}
We present multivariateGPT, a unified architecture for transformer-based modeling to discrete and numeric data. We show how multivariate timeseries data can be decomposed into an autoregressive prediction task and provide an embedding method and likelihood-based loss function for training a transformer model. Experiments on simple physical systems highlight that this approach enables generalization that is not apparent in discrete methods and is more sample efficient, reaching high precision in trajectory reconstruction in fewer iterations than discrete models. Furthermore, vocabularies are smaller because no discretization is necessary. Experiments on real-world clinical data show improved performance over state of the art models of clinical timeseries data with higher accuracy and better calibration. Additionally, mutlivariateGPT enables new types of predictions, namely predicting both the timing and value of observations. This is critical in irregular and informatively sampled data sets where the passage of time itself is critical.

This approach has broad potential in clinical settings where it could be used to train a foundation model on entire Electronic Health Record (EHR) databases. This could have broad impacts with improved prediction across a range of clinical tasks to improve decision making and resource utilization either through zero-shot prediction from Monte Carlo sampling of future trajectories or fine-tuning for specific tasks. The sequence representations could be used for patient search, matching, or phenotyping. This work has impact outside of medicine as it provides a framework for retrogressively modeling any database composed of categorical and numeric data which can be converted into class-value tuples.

\textbf{Limitations:} Limitations of this method include the current choice of Gaussian parameterization for value estimation. Although predictions are well calibrated using this parameterization on current data sets, an important direction for future work will be integrating different distributions that may better capture time, count, or ordinal data. This model also inherits limitations of transformer-based autoregressive modeling including interpretability of predictions and memory scaling with longer context windows. As this approach leverages the same architecture used in language models, we expect these limitations to improve with advancements in the field.

\textbf{Future Work:} Autoregressive image and language models show scaling of performance with model and dataset size which we hope to assess for this model. Not all numeric values are well represented by a continuous value estimated from a normal distribution no matter how well mean and variance are conditioned, such as non-negative, count, or ordinal data. We are working to incorporate likelihood based loss functions to specifically address these data types. We envision a mapping between common database data types (character, floating point, integer) and the embedding/loss functions proposed here. In this work we focus on low-dimensional numeric data (scalars across many categories), but we envision future work to incorporate additional data modalities such as speech and image data through established methods.

\section{Broader Impact}
Our work extends efforts for creating token-based autoregressive models to mixed categorical and numeric data streams using a single model architecture and likelihood based loss. This method is demonstrated to enable numeric generalization which was not seen in discrete token approaches and has state-of-the-art predictive performance. Timeseries prediction has numerous benefits in clinical applications, from identifying high-risk patients before critical events to optimizing resource allocation. In addition, sequence representations created by these models could be used for improved patient search or matching, enabling better clinical trial studies or recruitment. These benefits come with potential risks, including inaccurate predictions and propagation of biases in training data. If used improperly, this could lead to over- or under-treatment due to false positive or negative predictions.

While this work focuses on clinical applications, our method is flexible and can be used to model any data source that can be flattened into sequences of class-value tuples (for example by wide to long tabular data conversion). Future work will include application of this work to time-series data from other domains with similar risks of predictive errors in other high-risk use cases such as credit card fraud detection, sales forecasting, and job scheduling in information systems.

\begin{ack}
\section*{Acknowledgments}

AJL receives funding through the CTSA Grant UL1 TR001863 from the National Center for Advancing Translational Science (NCATS), a component of the National Institutes of Health (NIH). This publication’s contents are solely the responsibility of the authors and do not necessarily represent the official views of NIH.

\end{ack}

\bibliographystyle{abbrvnat} 
\bibliography{bibliography}


\appendix

\section{Appendix: Data Set and Experimental Details}

All code is available in the supplementary material and in the anonymized repository here: \url{https://anonymous.4open.science/r/multivariateGPT_anon-4ED4/README.md}. For all discrete models, bins are evenly spaced quantiles with resolution specified by the number of bins.

\subsection{Oscillator Data}
\label{appendix:oscillator}

\begin{figure*}[h]
    \centering
    \includegraphics[width=1.0\textwidth]{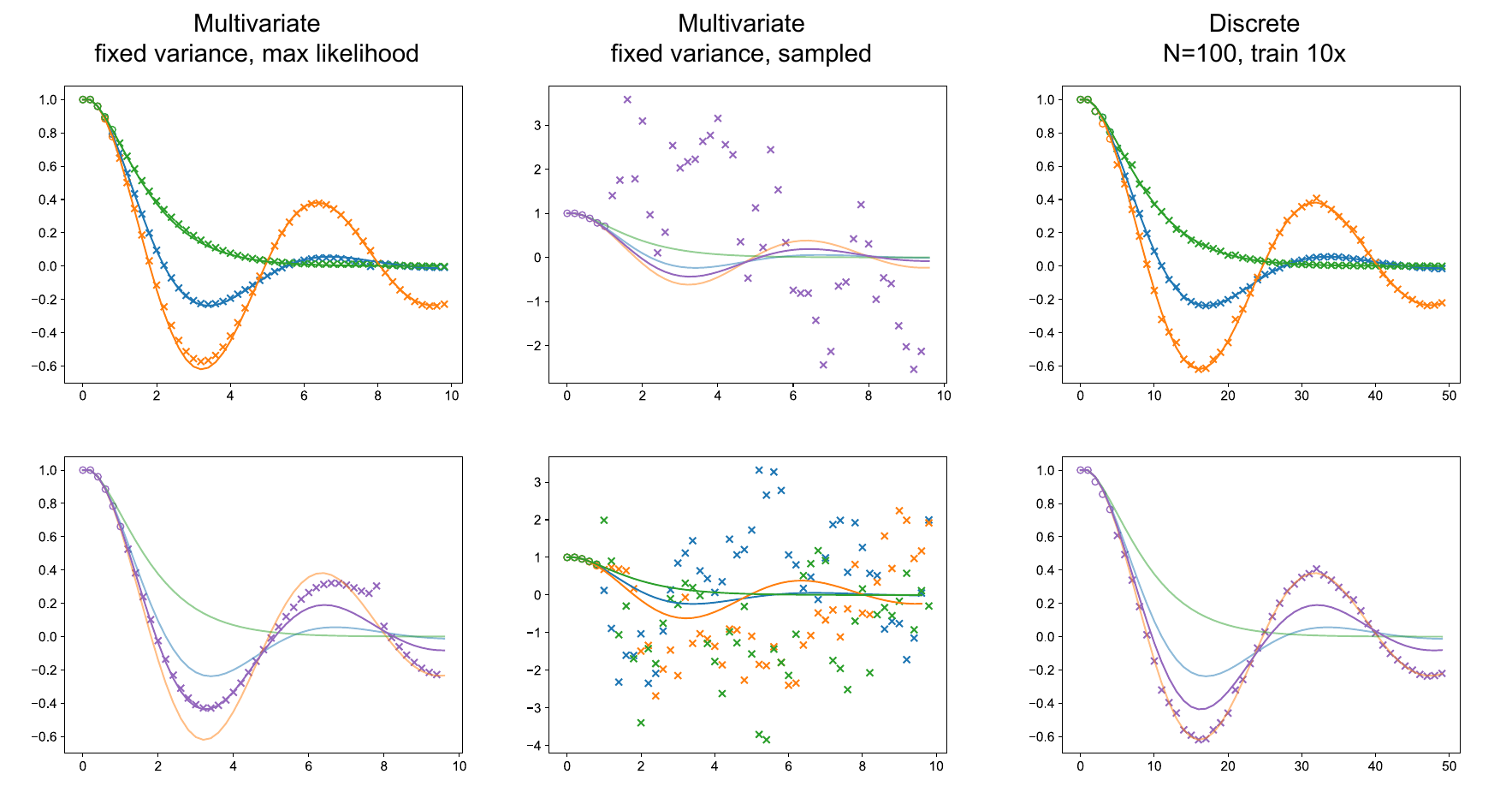}
    \caption{Additional simple harmonic oscillator experiments. Rows show task and columns show model type. The task in row 1 is reconstruction of training trajectories from seed points. The task in row 2 is generalizing to a trajectory that was not part of the training data. The multivariate model where variance is fixed and sampling is done by selecting the $(c,v)$ tuple with maximum likelihood shows similar performance on trajectories in training data is similar, but generalization fails (column 1). The multivariate model where variance is fixed an sampling is performed instead of selection of the value with maximum likelihood does not have enough numeric precision to fit trajectories (default variance is 1). The discrete model with 100 bins trained from 10 times longer than the multivariate model fits training data but fails to generalize (column 3)}
    \label{fig:oscillator_sup}
\end{figure*}

Ablation studies for learned variance and sampling as well as additional training durations for discrete models are shown in Figure \ref{fig:oscillator_sup}.

\subsection{MIMIC-IV Electrocardiogram}

\paragraph{Data Set:}
The MIMIC-IV electrocardiogram data set consisted of 89,335 electrocardiogram records with patient age, gender, and voltage time series for each of the 12 standard electrocardiogram leads (I, II, III, aVF, aVR, aVL, V1, V2, V3, V4, V5). The entire data set was divided into train, validation and test splits with 78,141, 3,185, and 8,009 records in each, respectively. The raw data was collected at 500 Hz and down-sampled to 100 Hz. A $4^{th}$ order Butterworth bandpass filter with a frequency range of 0.5 to 20 Hz was applied to each time series to reduce drift and high-frequency noise.
\paragraph{Model and Training Details:}
Model specifications are shown below in Table \ref{tab:model_configs_ecg}.
Models were trained for 20,000 iterations or until early stopping criteria were met. Warm-up and cosine decay for learning rate were used. The AdamW optimizer was used. 

\begin{table}[h!]
\centering
\setlength{\tabcolsep}{3.5pt}  
\begin{tabular}{@{}lrrrrrrrrr@{}}  
\hline
\multicolumn{1}{c}{\textbf{Model}} & 
\multicolumn{1}{c}{\textbf{Vocab}} & 
\multicolumn{1}{c}{\textbf{Param}} & 
\multicolumn{1}{c}{\textbf{n\_embd}} & 
\multicolumn{1}{c}{\textbf{n\_head}} & 
\multicolumn{1}{c}{\textbf{n\_layer}} & 
\multicolumn{1}{c}{\textbf{LR}} & 
\multicolumn{1}{c}{\textbf{Batch}} & 
\multicolumn{1}{c}{\textbf{Context}} & 
\multicolumn{1}{c}{\textbf{Steps}} \\
\hline
Discrete n=50  & 466 & 32.8M & 512 & 64 & 10 & $5{\times}10^{-4}$ & 65k & 1600 & 11200 \\
Discrete n=100 & 880 & 33.2M & 512 & 64 & 10 & $5{\times}10^{-4}$ & 65k & 1600 & 11200 \\
Numeric        & 12  & 31.6M & 512 & 64 & 10 & $5{\times}10^{-4}$ & 65k & 1600 & 16500 \\
\hline
\end{tabular}
\caption{Model configurations. Variables are as follows n\_embd: embedding dimension, n\_head: number of self-attention heads, LR: max learning rate, Batch: full batch size in tokens, Context: size of context window in tokens, Steps: number of training steps until early stopping criteria met or max training steps met.}
\label{tab:model_configs_ecg}
\end{table}

An example of lead reconstruction is shown below in Fig. \ref{fig:lead reconstruction}.
\begin{figure*}[h]
    \centering
    \includegraphics[width=1.0\textwidth]{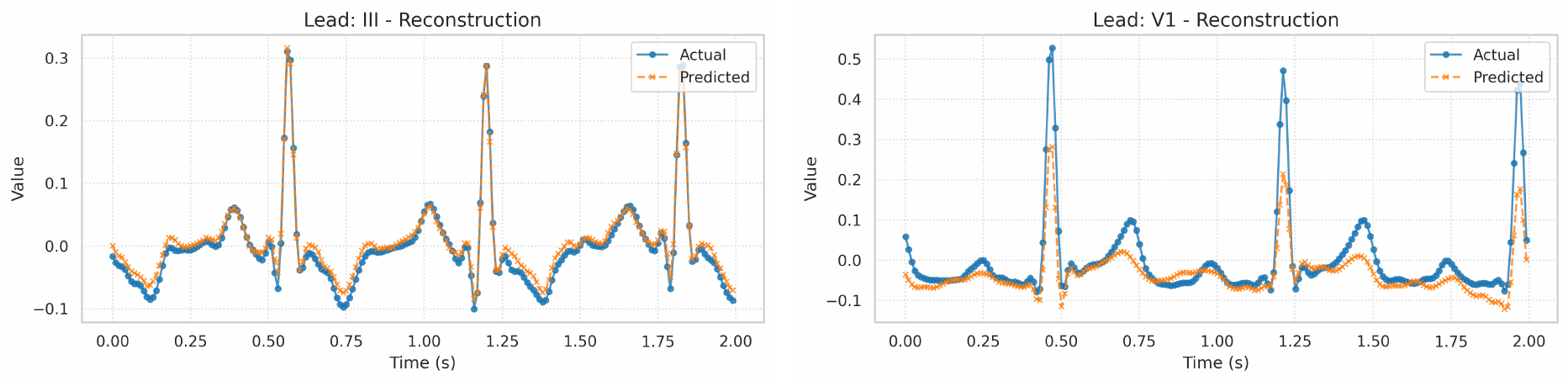}
    \caption{Example lead reconstructions of a limb lead (III) and precordial lead (V2).}
    \label{fig:lead reconstruction}
\end{figure*}

\subsection{eICU}

\paragraph{Model and Training Details:}

The following details the different model specifications and hyperparameters for training models on the eICU data (Table \ref{tab:model_configs_eicu}).

\begin{table}[h!]
\centering
\setlength{\tabcolsep}{3pt} 
\begin{tabular}{@{}lrrrrrrrrr@{}} 
\hline
\multicolumn{1}{c}{\textbf{Model}} & 
\multicolumn{1}{c}{\textbf{Vocab}} & 
\multicolumn{1}{c}{\textbf{Param}} & 
\multicolumn{1}{c}{\textbf{n\_embd}} & 
\multicolumn{1}{c}{\textbf{n\_head}} & 
\multicolumn{1}{c}{\textbf{n\_layer}} & 
\multicolumn{1}{c}{\textbf{LR}} & 
\multicolumn{1}{c}{\textbf{Batch}} & 
\multicolumn{1}{c}{\textbf{Context}} & 
\multicolumn{1}{c}{\textbf{Steps}} \\
\hline
Discrete n=10 & 65 & 25.5M & 512 & 8 & 8 & $1\times10^{-3}$ & 8192 & 512 & 5000 \\
Discrete n=50 & 252& 25.7M & 512 & 8 & 8 & $1\times10^{-3}$ & 8192 & 512 & 5000 \\
Numeric      & 14 & 25.4M & 512 & 8 & 8 & $1\times10^{-3}$ & 8192 & 512 & 5000 \\
\hline
\end{tabular}
\caption{Model configurations. Variables are as follows n\_embd: embedding dimension, n\_head: number of self-attention heads, LR: max learning rate, Batch Size: full batch size in tokens, Context: size of context window in tokens, Steps: number of training steps until early stopping criteria met or max training steps met.}
\label{tab:model_configs_eicu}
\end{table}

\paragraph{Comparison Model Details:} For TFM-ODE we used the implementation that was published for this data set in \citep{zhang2024trajectory}. For large-language model baseline, we used gemma 1.1-7b \citep{team2024gemma}. The following prompt was used: "You are tasked with predicting future values of heart rate and mean arterial pressure. the input data is: <data in csv format> give your best estimate to what hr and map will be at the following time: <times for estimation> and The norepi\_inf at these times is: <norepi infusion rates> use csv output with the following columns: time, hr, map."
\paragraph{Evaluation Details:}

To effectively combine MAP and heart rate for evaluation, we applied MinMaxScaler to normalize them to a consistent range. We fitted and transformed the ground truth values for MAP and HR separately and then applied the same transformation to the predicted values. MAP and HR were treated as a 2D array, and the MSE was calculated between the predictions and the ground truth.

To quantify performance variation associated with model training stability, we trained each model 3 times from a different random initialization and quantified variation in MSE (Table \ref{tab:multi_run_mse_results}).

\begin{table}[ht]
\centering
\small
\begin{tabular}{lcccc}
\toprule
\textbf{Model} & \textbf{3h} & \textbf{6h} & \textbf{9h} & \textbf{12h} \\
\midrule
Discrete (n=10) & 0.1160$\pm$0.0265 & 0.1084$\pm$0.0223 & 0.1063$\pm$0.0216 & 0.1017$\pm$0.0191 \\
Discrete (n=50) & 0.0317$\pm$0.0087 & 0.0296$\pm$0.0100 & 0.0291$\pm$0.0101 & 0.0274$\pm$0.0111 \\
Multivariate GPT & \textbf{0.0125$\pm$0.0006} & \textbf{0.0113$\pm$0.0007} & \textbf{0.0106$\pm$0.0011} & \textbf{0.0097$\pm$0.0008} \\
\bottomrule
\end{tabular}
\caption{Mean Squared Error (± Standard Deviation) for MAP and Heart Rate from various seed lengths for each model. Error intervals derived from multiple training runs.}
\label{tab:multi_run_mse_results}
\end{table}

\subsection{Physionet ICU Data}

The Physionet ICU data consisted of 65,155 unique records for patients admitted to the ICU. Data include demographics, vital signs, and lab values totaling 36 unique categorical and numeric measurement classes \citep{moor2021early}. Numeric classes were z-score normalized except for SpO2 which used a logistic scaling due to a ceiling effect where values cannot be greater than 100\% and are commonly at or near this value.

Model specifications are shown below in Table \ref{tab:model_configs_physionet}. Models were trained using a warm up period of approximately 5\% of the data followed by cosine learning rate decay. The AdamW optimizer was used. 

\begin{table}[h!]
\centering
\setlength{\tabcolsep}{3pt} 
\begin{tabular}{@{}lrrrrrrrrr@{}} 
\hline
\multicolumn{1}{c}{\textbf{Model}} & 
\multicolumn{1}{c}{\textbf{Vocab}} & 
\multicolumn{1}{c}{\textbf{Param}} & 
\multicolumn{1}{c}{\textbf{n\_embd}} & 
\multicolumn{1}{c}{\textbf{n\_head}} & 
\multicolumn{1}{c}{\textbf{n\_layer}} & 
\multicolumn{1}{c}{\textbf{LR}} & 
\multicolumn{1}{c}{\textbf{Batch}} & 
\multicolumn{1}{c}{\textbf{Context}} & 
\multicolumn{1}{c}{\textbf{Steps}} \\
\hline
Discrete n=10 & 372 & 0.91M & 128 & 4 & 4 & $1\times10^{-3}$ & 16384 & 512 & 10000 \\
Discrete n=50 & 1120& 1.1M & 128 & 4 & 4 & $1\times10^{-3}$ & 16384 & 256 & 10000 \\
Numeric      & 39 & 0.85M & 128 & 4 & 4 & $1\times10^{-3}$ & 16384 & 256 & 10000 \\
\hline
\end{tabular}
\caption{Model configurations. Variables are as follows n\_embd: embedding dimension, n\_head: number of self-attention heads, LR: max learning rate, Batch Size: full batch size in tokens, Context: size of context window in tokens, Steps: number of training steps.}
\label{tab:model_configs_physionet}
\end{table}

Due to the number of potential trajectories with 36 different values, we chose to assess the next class accuracy and next value prediction MSE.

\end{document}